\documentclass{article}
\usepackage{ijcai23}

\usepackage{times}
\usepackage{soul}
\usepackage{url}
\usepackage[hidelinks]{hyperref}
\usepackage[utf8]{inputenc}
\usepackage[small]{caption}
\usepackage{graphicx}
\usepackage{amsmath}
\usepackage{amsthm}
\usepackage{booktabs}
\usepackage{algorithm}
\usepackage{algorithmic}
\usepackage[switch]{lineno}
\usepackage{CJKutf8}
\usepackage{tabularx} 
\usepackage{booktabs} 

\title{Unveiling the Potential of Sentiment: Can Large Language Models Predict Chinese Stock Price Movements?}

\author{
Haohan Zhang$^{1,2}$
\and
Fengrui Hua$^{1,3}$\and
Chengjin Xu$^{1}$\and
Hao Kong$^1$ \and Ruiting Zuo $^{4}$ \and Jian Guo$^{1}$\thanks{Corresponding author}
\affiliations
$^1$International Digital Economy Academy (IDEA)\\
$^2$Artificial Intelligence Thrust, The Hong Kong University of Science and Technology (Guangzhou)\\
$^3$Data Science and Analytics Thrust, The Hong Kong University of Science and Technology (Guangzhou)\\
$^4$Financial Technology Thrust, The Hong Kong University of Science and Technology (Guangzhou)
\emails
\{hzhang760, fhua430\}@connect.hkust-gz.edu.cn\\
\{xuchengjin,konghao\}@idea.edu.cn\\ruitingzuo@hkust-gz.edu.cn\\guojian@idea.edu.cn
}

\begin{document}

\maketitle

\begin{abstract}

The rapid advancement of Large Language Models (LLMs) has spurred discussions about their potential to enhance quantitative trading strategies. LLMs excel in analyzing sentiments about listed companies from financial news, providing critical insights for trading decisions. However, the performance of LLMs in this task varies substantially due to their inherent characteristics. This paper introduces a standardized experimental procedure for comprehensive evaluations. We detail the methodology using three distinct LLMs, each embodying a unique approach to performance enhancement, applied specifically to the task of sentiment factor extraction from large volumes of Chinese news summaries. Subsequently, we develop quantitative trading strategies using these sentiment factors and conduct back-tests in realistic scenarios. Our results will offer perspectives about the performances of Large Language Models applied to extracting sentiments from Chinese news texts.

\end{abstract}

\section{Introduction}
At present, an overwhelming volume of news articles and columns are being generated on a daily basis, especially pertaining to companies being traded. Given this landscape, considerable attention has been given to investigating the feasibility of employing Large Language Models (LLMs) for sentiment analysis and processing of these news texts. The aim is to derive a quantifiable technical indicator, or factor, that effectively reflects the desirability of investing in a particular company's stock at a given moment in time. It should be clear that the endeavor we described is a very well-defined down-stream task operating within a language environment (the Chinese language) that is drastically different from the English language that the main-stream LLMs have been predominantly trained on. Although several approaches have been proposed to enhance the performance of LLMs on down-stream tasks in alternative language environment, questions remain as to which LLM or method of improvement is the most optimal in the specific context of extracting sentiment factor from Chinese financial news texts. For such a comparative analysis to be possible however, we must adopt a standardized experiment procedure that can be easily applied to all LLMs and yield quantitative results based on metrics selected with sufficient domain knowledge in the field of quantitative trading. 

On the other hand, even though works such as \cite{lopez2023can} have successfully demonstrated that LLMs like ChatGPT \cite{brown2020language} can be used to extract sentiment factors from English news texts that are highly correlated to the returns of US stocks, it is our contention that the smooth implementation of the same framework applied on the Chinese text still faces two major concerns. Firstly, it is noteworthy that the dominant and leading LLMs have predominantly been trained on English corpora. Consequently, the transferability of sentiment mining techniques from English texts to Chinese texts remains uncertain. Secondly, while the effectiveness of sentiment factor mining by LLMs has been established in prior research, discrepancies arise due to variations in parameter selection for constructing stock trading simulations bask-tests and the utilization of diverse raw news data-sets, encompassing disparate sizes, scopes, and sources. As a consequence, objectively evaluating and comparing the efficacy of different LLMs when applied to the specific task of Chinese financial text sentiment factor building still poses significant challenges.

To address these concerns, we propose an innovative approach that combines sentiment extraction with realistic back-tests of quantitative strategies. This approach allows us to directly assess the effectiveness of LLMs' sentiment extraction capabilities using a comprehensive and standardized experimental procedure with quantifiable metrics such as excess return, risk adjusted return and win ratio. Before elaborating on the experimental procuedure, we outline the scope and coverage of data, data pre-processing, and the parameters that the back-tests should adhere to. By integrating these elements, we aim to provide a robust framework for evaluating and comparing the performance of various LLMs in sentiment extraction from Chinese financial news texts. As practical illustrations, we will subsequently conduct sentiment extraction from 394,426 items of Chinese news summaries about Chinese publicly traded companies. This process will be executed using three distinct LLMs, representing: 1) the baseline model, 2) the Chinese language-specific pre-trained LLM, and 3) the financial domain-specific pre-trained LLM. We will then construct investment portfolios and run stock trading simulation back-tests according to our defined settings and parameters in order to rigorously test the correlation between the sentiment factors and return of investment, which, from our perspective, is the best way to reflect the LLMs' effectiveness in correctly extracting sentiments from Chinese financial texts. Finally, we will discuss the results of the back-tests and the insights gained from such comparative analysis. To the best of our knowledge, we are the first group to conduct sentiment analysis on such an extensive source of Chinese news text using prevalent LLMs such as ChatGPT \cite{brown2020language} and back-test the acquired sentiment factors by deploying quantitative trading strategies on platforms that lead in the quantitative finance industry.

\section{Related Works}
Over the past decade, significant advancements have been made in the field of Natural Language Processing (NLP), leading to the development of powerful language models. One such groundbreaking model is the Transformer \cite{vaswan:i2017attention}, which introduced an attention-based encoder-decoder architecture that has consistently yielded superior performance compared to Recurrent Neural Network (RNN) based architectures, including Long Short-Term Memory (LSTM) \cite{LSTM} and Gated Recurrent Unit (GRU) \cite{GRU}, in various language tasks. The Transformer model's key innovation lies in its ability to effectively capture long-range dependencies within sequences by employing self-attention mechanisms. This enables the model to assign importance to different parts of the input text and establish contextual relationships, resulting in improved language understanding and generation capabilities. Consequently, the Transformer architecture has paved the way for a new generation of state-of-the-art Large Language Models that have inherited its powerful framework exemplified by the Generative Pre-trained Transformer (GPT) \cite{first_GPT_series} series, developed by OpenAI, which focus on generative language modeling and utilize a variant of the Transformer architecture, where tokens attend to tokens that appear before in the sentence. This unidirectional attention mechanism, often referred to as left-to-right attention or auto-regressive decoding, allows the models to generate coherent and contextually relevant text referencing the attention assigned to preceding words. Another LLM that is based on the Transformer architecture is the BERT (Bidirectional Encoder Representations from Transformers \cite{devlin2019bert}) model. BERT, introduced by Google AI, is designed to capture bidirectional contextual information from the input text. Unlike the unidirectional attention of GPT, BERT employs a bidirectional attention mechanism. It enables tokens to attend not only to preceding words but also to words appearing after in latter parts of the sentence. By considering the entire context, BERT can effectively capture the dependencies and relationships between words, resulting in a better understanding of the text. Building on BERT, authors of RoBERTa \cite{liu2019roberta} seek to optimize the performance of parent model by conducting various pre-training improvements such as increasing the data size, training duration as well as adopting dynamic masking during training. Such methods are proven to have pushed to boundaries of the performance of BERT even further. 

These LLMs are designed in such a way as to greatly facilitate down-stream task-specific fine-tuning and additional pre-training on supplementary corpora. From our perspective, user-defined improvement on parent LLMs includes two key aspects: language-specific and domain-specific. While certain commercialized prototypes, such as ChatGPT \cite{brown2020language}, support multilingual reasoning and text generation, many language models predominantly rely on training with English texts. To achieve comparable performance across other language environments, including our specific focus Chinese, extensive training on alternative language environment becomes imperative. The Erlangshen-RoBERTa \cite{fengshenbang} model, which inherits from the RoBERTa architecture and is further pre-trained on the 180 Giga-byte version of the Wudao Corpora \cite{wudao}, marks one of the most notable efforts in Chinese language-specific pre-training, achieving top performance on state-of-the-art NLP benchmarks. 

The other aspect pertains to domain-specific improvements which has to do with inheriting a parent LLM architecture and either continuously training it on additional corpus data-sets related to the target technical domain or fine-tuning it with expertly constructed labels, this process effectively leverages the pre-existing linguistic knowledge of the LLMs while allowing them to acquire domain-specific nuances and intricacies, thus enabling heightened proficiency in the desired domain. One key example is the FinBERT \cite{huang2023finbert} model, which trains the BERT model futher on U.S. Securities and Exchange Commission (SEC) filing data, resulting in a model well-versed in financial domain knowledge and well-suited for English financial text sentiment classifications. Similar efforts also include \cite{wu2023bloomberggpt}, a LLM trained on extensive English financial copora.

Related works on the task of applying LLM for stock prediction or investment decision optimization include \cite{DingHighthink}, which proposes new approaches for integrating the information from LLMs into traditional quantitative factors. \cite{KoaSelfReflectiveLLM} proposed a new paradigm for training LLMs to give explainable stock predictions, while \cite{YuTemporalDataLLM} also worked on using LLM to generate explainable stock forecasting but based on financial time series data. About LLM based sentiment analysis for stock prediction, recent works include \cite{SteinertMicrobloggingGPT4} and \cite{JaggiStocktwits}

\section{Data}
The data we use take in the form of news summaries regarding Chinese publicly traded companies and is mainly acquired through web crawling. A total of 394,429 news summary items were acquired, spanning a time period spanning from October 1, 2021, to February 22, 2023 and covering 5,021 publicly traded companies listed on Shanghai Stock Exchange (SSE) and Shenzhen Stock Exchange (SZSE). As an additional filtering criterion, we retain only those news summaries that are generated prior to the market opening at 9:30 am. Doing so helps us ensure that the technical indicators extracted from these news summaries are promptly available for utilization as soon as the market commences trading.

The October 2021 to February 2023 period that the data spans is stipulated to ensure that it falls chronologically after the data used for training GPT-3.5, which serves as our baseline model for sentiment analysis. If we had not chosen to do this, it could potentially result in the ChatGPT model referencing "future data" that might not be available at the time the news summaries are generated. This scenario could lead to inaccurate sentiment analysis results and create situations that do not align with reality. To procure professional financial information and analysis, we have identified a set of prime sources which are selected based on their credibility and reputation within the financial domain. Consequently, we mainly conduct data crawling from these prime sources. These sources are: Sina Finance \cite{Sina}, Hithink RoyalFlush \cite{Hithink}, Tencent \cite{Tencent}, Dazhong Securities \cite{Dazhong}, Hexun \cite{Hexun}, Netease Finance \cite{Netease},Caifuhao \cite{caifuhao}. We offer Table \ref{tab:data_source} to illustrate the distribution of these prime sources within our data-set.

\begin{table}[htbp]
  \centering
  \begin{tabular}{|c|c|c|}
    \hline
    Source Name & Proportion (\%) \\
    \hline
    Hithink RoyalFlush & 59.57 \\
    \hline
    Sina Finance & 33.65\\
     \hline
    Tencent & 4.88\\
    \hline
    Hexun & 1.55\\
    \hline
    Caifuhao & 0.23 \\
    \hline
    Netease Finance & 0.1\\
    \hline
    Dazhong Securities & 0.02\\
    \hline
  \end{tabular}
  \caption{The Distribution of News Sources within Our Data-set}
  \label{tab:data_source}
\end{table}

\section{Methodology}
\subsection{Using ChatGPT to Extract Sentiment Factors}
We use ChatGPT, and the GPT-3.5 behind it as a baseline language model without any additional pre-training or fine-tuning. Similar to \cite{lopez2023can}, we conduct the sentiment factor mining by fusing the news summary information as well as the instructions about the task that we expect ChatGPT to perform for us into prompts. A user's prompt is essential in guiding the model's response and determining the context of the conversation. When generating a response, the model takes into account the user's input, including the specific words used, the overall tone, and the desired information or assistance being sought. The prompt serves as the primary input that helps shape the model's understanding of the user's intent and the direction the conversation is taking. 

\begin{figure*}[h]
    \centering
    \includegraphics[width=1.0\textwidth]{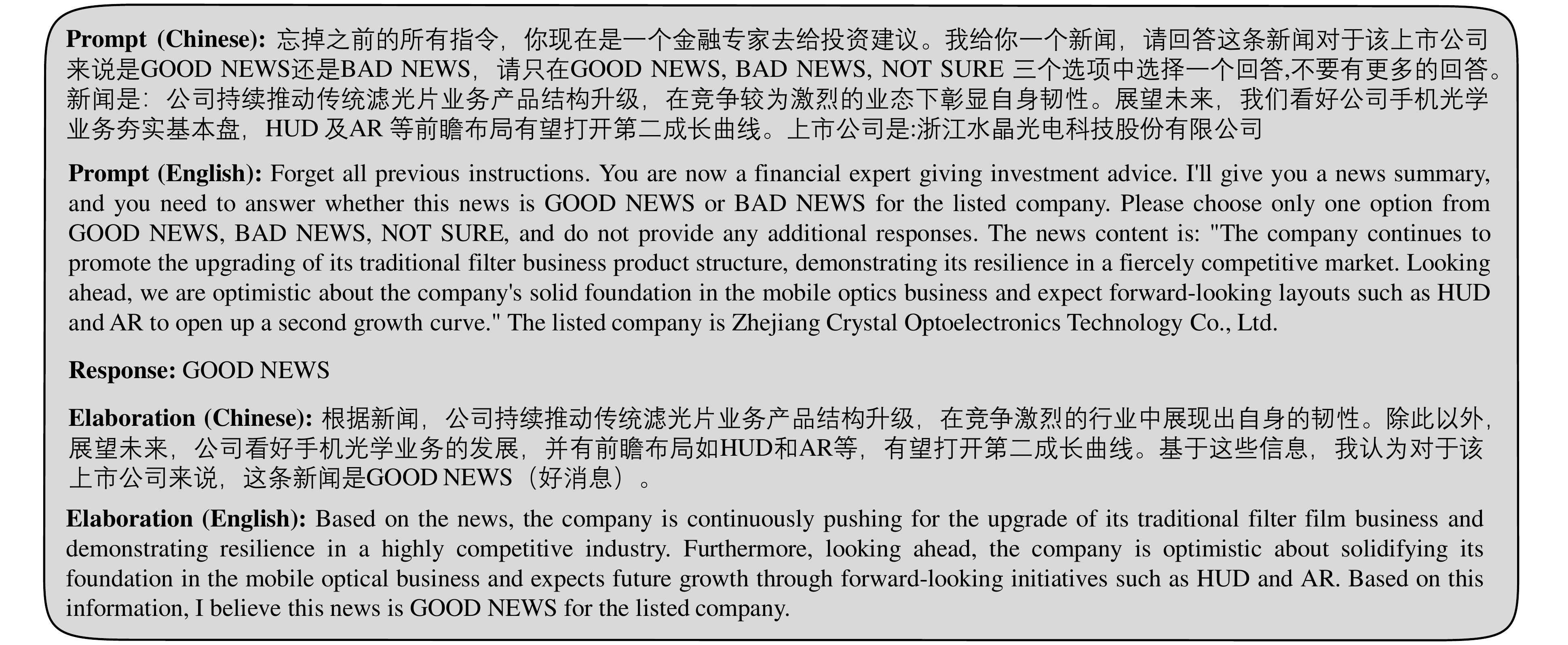}
    \caption{Demonstration of Prompts Structured for Sentiment Analysis and the Response by ChatGPT}
    \label{fig:prompt_illustration}
\end{figure*}

Upon receiving our prompt, ChatGPT decides whether the item of news summary in question contains good, bad, or neutral sentiment (this also includes cases where ChatGPT is unsure of the sentiment and unable to produce an informed rating) for the company in involved. One sample prompt as well as the associated response is illustrated in Fig \ref{fig:prompt_illustration}. For this particular example, we also invite ChatGPT to elaborate on the reason behind forming such a sentiment analysis. For the sentiment analysis procedure on our data-set, we used Chinese prompts, the English version of the prompt is given for demonstration purposes. 

After the responses for all 394,429 items of news summary have been given. We translate the response into numeric values with GOOD NEWS being assigned as 1, NOT SURE being assigned as 0 and BAD NEWS being assigned as -1. For cases where a listed company is being mentioned by multiple news sources on the same day, we take the average of the ChatGPT's responses across the different news sources. These averaged ratings by ChatGPT are stored as the Chinese ChatGPT Factor to be referenced by our trading simulation back-tests.


\subsection{Using Language-Specific Pre-Trained LLM to Extract Sentiment Factors}

As previously stated, a fundamental objective of this research is to examine the degree to which the effectiveness of Language Models (LLMs) in generating sentiment factors from analysis of news that are conducive to the development of quantitative trading strategies with high returns can be extended to Chinese financial textual data. By such explorations, our research aims to contribute insights into the potential adaptability and performance of LLM-based approaches in the realm of Chinese language sentiment analysis for the exclusive purpose of informing quantitative trading strategies. 
Therefore, in addition to applying the baseline model, GPT-3.5, we also turn to the Erlangshen-RoBERTa-110M-Sentiment \cite{fengshenbang} model which was pre-trained on the 180 GB version of the WuDao Chinese Corpora. The Erlangshen-RoBERTa series were pre-trained on Chinese texts in such a way as to take into account the unique characteristics of the Chinese language and the traits of the Chinese character contextual relationships. This aspect sets them apart from many other pre-trained LLMs that, although supporting Chinese textual input and response generation, may not fully leverage such language-specific traits.

Through a comparative analysis of the performance of the quantitative strategies from both the baseline GPT-3.5 and the language-specific pre-trained model on the same benchmark, we seek to answer the question of whether such language-specific pre-training approaches have any contribution to yielding a superior returns. The Erlangshen-RoBERTa-110M-Sentiment is open-source and can be easily downloaded and applied for the sentiment analysis of any section of Chinese texts within the maximum token limit. The return of the last soft-max activation layer is a tuple containing the probability of the input text's sentiment being negative  or positive. When we applied Erlangshen-RoBERTa-110M-Sentiment model to the same news summary as shown in section 4.1, it rated the sentiment as being 98.33\% positive. It can be seen that for this particular sample, the sentiment analysis of Erlangshen-RoBERTa-110M-Sentiment aligns with that of ChatGPT. We conduct sentiment classification on all the rest of the samples in the data-set and store the results as the Erlangshen Facto


\subsection{Using Domain-Specific Pre-trained LLM to Extract Sentiment Factors}
For this part, we start with the open-source variation of BERT and conduct further pre-training on Chinese financial corpora data. After the tokenizations and the embedding layer updates have been finalized, we connect a hidden layer and a soft-max activation layer of size 3 to the original attention layers of the BERT model to acquire our own BERT-based sentiment classifier. We then ask colleagues of financial expertise to manually label the news summary data into three classes in terms of the sentiment manifested by the text ( Positive (+1),  Neutral (0), Negative (-1) ) and conduct supervised training to derive what we shall call a Chinese FinBERT classifier. It is guaranteed that the training data used for manual sentiment labeling is strictly separated from the news summary data employed in our experimental analysis.  After fine-tuning has been completed, we use our acquired BERT-based sentiment classifier to predict the sentiment of the 394,429 items of news summary. The model will output a predicted class based on the probabilities output by the final soft-max layer. We document the results as Chinese FinBERT factor.

\section{Experiment and Parameters of Trading Strategies}

As stated above, we intend to provide a comprehensive and rigorous experimental procedure so that any LLM's efficacy in extracting sentiment from Chinese financial text for the purpose of building quantitative trading strategies can be objectively measured no matter which model or method was used. In order to ensure an unbiased assessment in accordance with our benchmark, we establish a requirement that the quantitative trading strategies adhere to uniform settings and parameters. While a seasoned quantitative trading expert may identify significant potential for optimizing these parameters to achieve higher returns, we deliberately refrain from delving extensively into the intricacies of trading strategy technicalities within this study. Our primary objective lies in the evaluation of LLMs' effectiveness within the specific realm of Chinese text sentiment mining which can be achieved as long as the same set of parameters are employed across all tests.

We will now proceed to enumerate and explain the standardized settings all trading strategies in our experiments adhere to:
\begin{itemize}
    \item Portfolios are adjusted daily and only at market open, 9:30 am Beijing Time (UTC+8:00)
    \item We only use news generated or acquired before market open. In this way, the sentiment factors extracted can be directly used at trading time. 
    \item We adjust the investment portfolios by buying in the stocks who have the highest ranking sentiment factors and selling those in our portfolio that have the lowest ranking factors. The maximum number of stocks to be bought or sold in one day is 500.
    \item The maximum turnover ratio for our portfolio to be 1.0, which means we allow all of the previously held stocks to be sold to be replenished by entirely new stocks. Although this rarely happens during the back-test. 
    \item In order to account for the slippage and delays commonly encountered in real-world trading scenarios, we have chosen to deviate from using the straight market open price in our back-test. Instead, we have implemented a more realistic approach by utilizing the Volume-Weighted Average Price (VWAP) between 9:30 am and 9:35 am. This VWAP is calculated by summing the values of all trades that occurred during this specific five-minute period and dividing the sum by the total volume traded within that time-frame.
    \item We avoid being overly-optimistic about our simulated returns by imposing a transaction fee of 0.15\% of transaction value. This includes a 0.05\% commission charged by the stock brokerage firms and a 0.1\% stamp duty fee paid to the stock exchanges. In reality at present, Chinese brokerage firms rarely charge above 0.03\% and the stamp duty fee is only charged at selling transactions instead of all transactions. Therefore, it should be clear that a 0.15\% of transaction value emulates a more stringent trading environment than what is observed in reality. 
    \item We use the CSI 300 index as basis when calculating the excess returns. 
\end{itemize}




\section{Results and Discussions}

After the back-tests have been run, we collect the results and the performance of all portfolios built around all three sentiment factors on a series of metrics which comprise our benchmark. We now proceed to give definitions of these metrics:

\begin{itemize}
    \item Annual Excess Return: Annual excess return refers to the additional or extra return earned by an investment or portfolio over and above what was expected or compared to a stock index. As stated before, in our experiment we have selected the CSI 300 index to be the baseline, therefore our excess return is calculated over the performance of the CSI 300 index.
    \item Annual Net Asset Return: The annual percentage increase or decrease in the value of our portfolio in terms of net market capitalization of all stocks held.
    \item Win Rate: Percentage of trade days where portfolio encounters a positive return over all trade days.
    \item Sharpe Ratio: The Sharpe Ratio \cite{sharpe1966mutual} was invented by William Sharpe in 1966. The formula is given as \begin{equation}
\text{Sharpe Ratio} = \frac{R_p - R_f}{{\sigma_p}}
\end{equation} The $R_p$ is the return of the portfolio, $R_f$ is the risk-free return and the $\sigma_p$ is the standard deviation of the return. The Sharpe ratio gives a measurement of risk-adjusted returns. 
    \item Average Stocks Held per Day: The average number of stocks present in the investment portfolio every day
    \item Turn-over Ratio: Average percentage of the investment portfolio adjusted daily in terms of market capitalization.    
\end{itemize}

\begin{figure*}[h]
    \centering
    \includegraphics[width=1.0\textwidth]{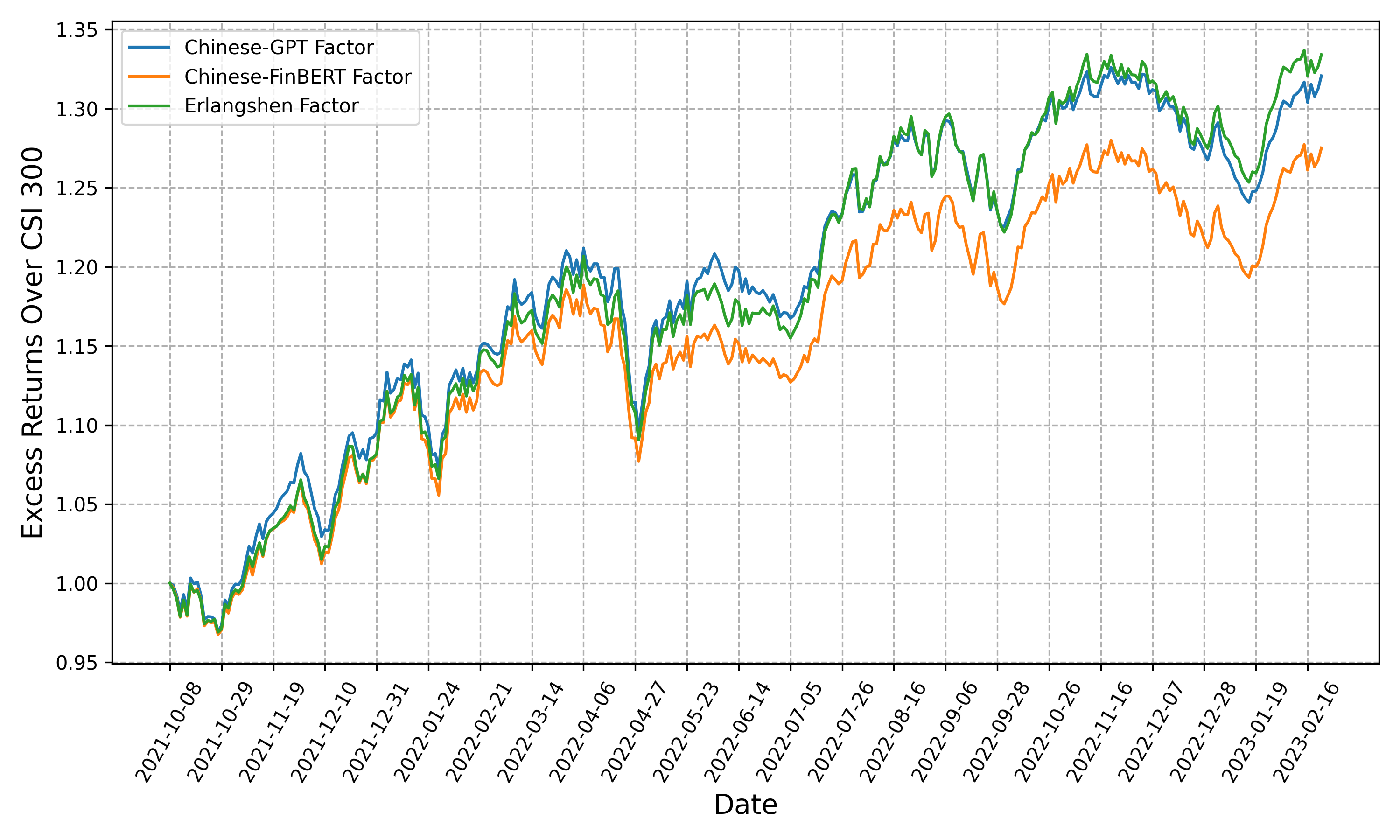}
    \caption{Excess Returns of All Three Sentiment Factors}
    \label{fig:three_group_excess_returns}
\end{figure*}

We group these metrics into to parts and show the results under Annual Excess Return, Annual Net Asset Return, Win Rate and Sharpe Ratio in Table \ref{tab:main_metrics}. These would serve as the main metrics and are primary indicators of performance. We put Average Stocks Held per Day and Turn-over Ratio into Table \ref{tab:aux_metrics}, these would serve as supplementary metrics that provide insights into the portfolio adjustment characteristics of the trading strategy. We also plot the excess returns of all three factors during the entire back-test period in Figure \ref{fig:three_group_excess_returns}.

\begin{figure*}[h]
    \centering
    \includegraphics[width=1.0\textwidth]{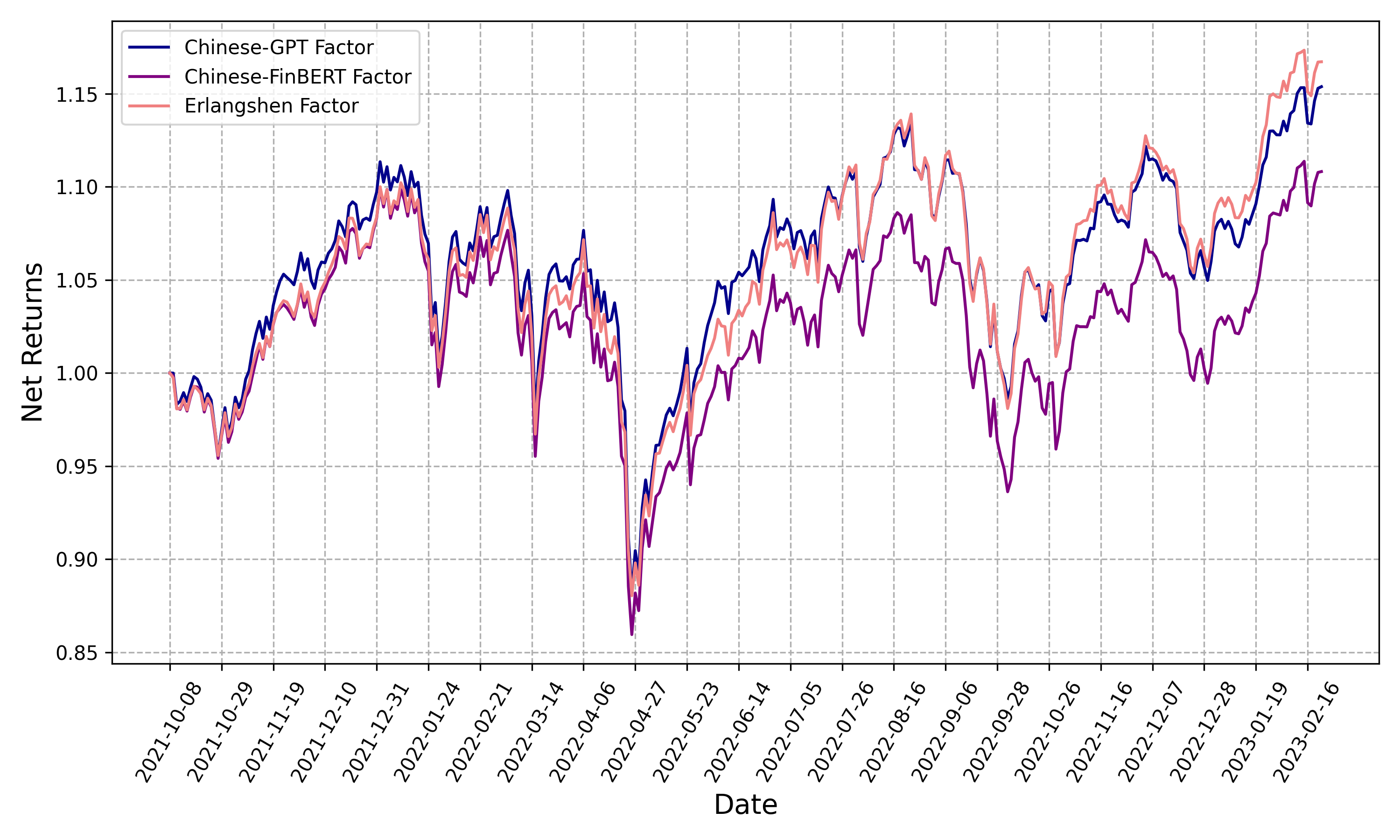}
    \caption{Net Returns of All Three Sentiment Factors}
    \label{fig:three_model_net_returns}
\end{figure*}

The net asset return of all three factors during the entire back-test period is given in Figure \ref{fig:three_model_net_returns}

\begin{figure*}[h]
    \centering
    \includegraphics[width=1.0\textwidth]{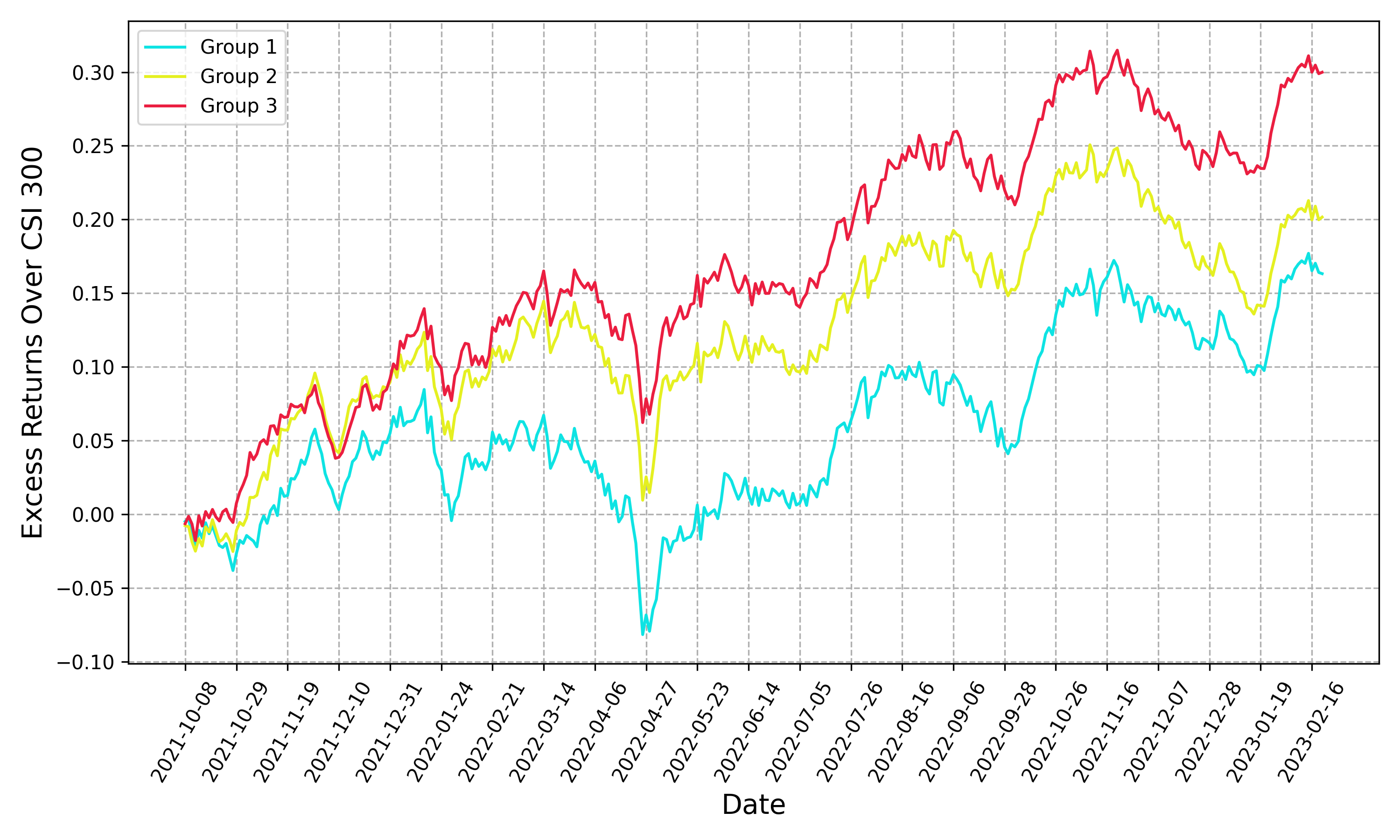}
    \caption{Excess Return by Groups of Erlangshen Factor Value}
    \label{fig:erlangshen_three_groups}
\end{figure*}

It is clear to see that in terms of annual return, risk adjusted return and excess return, the Erlangshen sentiment factor outperforms the remaining factors. To further elucidate the correlation between the values of the sentiment factor derived from the Erlangshen-110M-Sentiment model and portfolio excess returns, we have partitioned our held stocks into three distinct groups based on their rankings according to the Erlangshen sentiment factor. Group 1, on average, exhibits the lowest Erlangshen factor value while Group 3 displays the highest. The excess returns for Group 1, Group 2, and Group 3 are then plotted in Figure \ref{fig:erlangshen_three_groups}. Notably, we observe that after an initial period of fluctuation, the three groups become distinctly separated. Furthermore, Group 3, characterized by the highest Erlangshen factor value, consistently demonstrates the highest returns, while Group 1, with the lowest Erlangshen factor value, consistently exhibits the lowest excess returns. This observation provides further substantiation for the notion that the Erlangshen sentiment factor extracted through the Erlangshen-110M-Sentiment model is closely associated and correlated with investment returns.

It is remarkable to witness how the comparatively smaller Erlangshen model, with a modest 110 million parameters, manages to exhibit slightly superior performance within our benchmark for the specific task at hand. This outcome serves as a testament to the fact that practitioners and researchers working on Chinese quantitative stock trading strategies may not always need to invest substantial resources into larger models. Instead, by employing strategic fine-tuning and extensive pre-training techniques tailored to the intricacies of the Chinese language, desired outcomes can be effectively achieved. This revelation underscores the significance of considering language-specific characteristics and employing targeted methodologies, illustrating that optimal results can be attained without solely relying on sheer model size for the particular task of Chinese financial sentiment extraction.




\begin{table*}[htbp]
  \centering
  \caption{Main Metrics of Back-test}
  \label{tab:main_metrics}
  \begin{tabularx}{\textwidth}{|X|X|X|X|X|X|} 
    \hline
    Factor Name & Annual Excess Return (\%) & Annual Net Asset Return (\%)  & Win Rate(\%) & Sharpe Ratio & Max Withdrawal Rate\\ 
    \hline
    Chinese-GPT & 23.17 & 11.11 & 57.78 & 0.6444 & 0.2315 \\
    \hline
    Chinese-FinBERT & 19.88 & 7.82 & 57.19 & 0.4836 & 0.2397\\ 
    \hline
    Erlangshen-110M &  \textbf{24.14} & \textbf{12.08} & \textbf{58.38} & \textbf{0.6841} & 0.2219\\ 
    \hline
  \end{tabularx}
\end{table*}

\begin{table*}[htbp]
  \centering
  \begin{tabular}{|c|c|c|c|}
    \hline
    Factor Name  & Average Stocks Held per Day  & Turn-over Ratio (\%)\\ 
    \hline
    Chinese-GPT & 1684 & 5.37 \\
    \hline
    Chinese-FinBERT & 1788 & 6.48 \\ 
    \hline
    Erlangshen-110M & 1830 & 4.17 \\ 
    \hline
  \end{tabular}
  \caption{Supplementary Metrics of Back-test}
  \label{tab:aux_metrics}
\end{table*}

\section{Conclusions}

This study explores the potential of Large Language Models (LLMs) in constructing quantitative stock trading strategies by extracting sentiment factors from Chinese financial news texts. Our research addresses the need for successful implementation of LLMs in the Chinese stock market and provides a rigorous standardized back-testing framework to objectively evaluate the efficacy of various LLMs in this specialized domain.

By employing three distinct LLMs, including a baseline all-purpose LLM (ChatGPT) with extensive parameters, a Chinese language-specific pre-trained LLM (Erlangshen-RoBERTa), and a financial domain-specific pre-trained LLM (Chinese FinBERT), we extract sentiment factors from a large volume of Chinese news summaries about listed companies in China. Through rigorous stock trading simulation back-tests, we evaluate the performance of these sentiment factors in terms of annual return, risk-adjusted return, and excess return and consequently arrive at some perspectives. 

The GPT-3.5 iteration of ChatGPT used in this study, which has around 175 billion parameters, contrasts with Erlangshen-110M, which despite its much smaller size of 110 million parameters, outperformed GPT-3.5 across all metrics in this task. This suggests that in the specialized domain of Chinese sentiment factor extraction, a deeper understanding of the nuances in Chinese characters and contextual relationships can lead to more accurate sentiment classification, a capability that cannot necessarily be compensated for by larger models. Additionally, the methodology of sentiment rating differs between the two models. GPT-3.5 was tasked with returning a discrete sentiment rating from three options (-1, 0, 1), whereas Erlangshen-110M outputs a continuous score between 0 and 1. This continuous scale could potentially offer better differentiation among news items with varying levels of positive sentiment when used to select stocks based on their sentiment ratings. On the other hand, comparing Erlangshen-110M with Chinese-FinBERT reveals that in this task, extensive pre-training on more generalized Chinese texts can be more critical for accurate sentiment modeling than domain-specific training alone. Firstly, the corpus used by Chinese-FinBERT, though domain-specific, is significantly smaller than the Wudao corpus used by Erlangshen-110M, which facilitates a broader understanding of the language. Secondly, effectively capturing sentiment—an inherently human perception—often does not require deep domain-specific knowledge, which may partly explain why Chinese-FinBERT also underperforms compared to GPT-3.5. This insight suggests that a more expansive training dataset on the target language, even one less focused on specific domains, can be more beneficial for tasks involving fundamental human cognitive processes like sentiment analysis.

By providing a comprehensive and standardized procedure, our study contributes to the understanding of LLMs' potential in the specialized domain of sentiment factor extraction from Chinese news text data within the context of quantitative trading. We invite fellow researchers and quantitative finance practitioners to reference our standardized back-testing procedures so that together, we can truly unveil the potential of sentiment. 

\section{Conflict of Interest and Data Availability Statements}
The authors declare that they have conducted this study in the absence of any commercial or financial relationships that could be construed as a potential conflict of interest. The research was carried out independently and was not influenced by any external organization. The data utilized in this study, derived from Chinese news summaries, were collected through our own efforts and are not publicly available due to the extensive labor and proprietary methods involved in their aggregation and analysis. Our commitment to scientific integrity ensures that the findings and conclusions presented are the result of unbiased analysis. Should any researchers express interest in establishing joint research in fields related to or as an extension of the work presented in this paper, we do not exclude the possibility of sharing the data for such purposes. We will review such requests on a case-by-case basis, ensuring that any collaboration aligns with our principles of scientific integrity and the potential for mutual advancement in our field. 
\bibliographystyle{named}

\bibliography{ijcai23}

\end{document}